\pdfoutput=1

\documentclass[11pt]{article}

\usepackage{ACL2023}

\usepackage{times}
\usepackage{latexsym}
\usepackage{graphicx}
\usepackage{amsmath}
\usepackage{amssymb}
\usepackage{booktabs}
\usepackage{mathabx}
\usepackage{amsmath}
\usepackage{enumitem}
\usepackage{adjustbox}
\usepackage{listings}

\usepackage[T1]{fontenc}

\usepackage[utf8]{inputenc}

\usepackage{microtype}

\usepackage{inconsolata}

\title{KAFA: Rethinking Image Ad Understanding with Knowledge-Augmented Feature Adaptation of Vision-Language Models}

\author{Zhiwei Jia\thanks{\ \ \  Work done in part during an internship at Google. Correspondence to \texttt{zjia@eng.ucsd.edu}.}
 \\
  UC San Diego  \\
\And
  Pradyumna Narayana \\
  Google \\
\And
  Arjun R. Akula \\
  Google \\
\AND
  Garima Pruthi \\
  Google \\
\And
  Hao Su \\
  UC San Diego \\
\And
  Sugato Basu \\
  Google \\
\And
  Varun Jampani \\
  Google \\
}

\begin{document}
\maketitle
\begin{abstract}
   Image ad understanding is a crucial task with wide real-world applications. 
   Although highly challenging with the involvement of diverse atypical scenes, real-world entities, and reasoning over scene-texts, how to interpret image ads is relatively under-explored, especially in the era of foundational vision-language models (VLMs) featuring impressive generalizability and adaptability.
   In this paper, we perform the first empirical study of image ad understanding through the lens of pre-trained VLMs.
   We benchmark and reveal practical challenges in adapting these VLMs to image ad understanding.
   We propose a simple feature adaptation strategy to effectively fuse multimodal information for image ads and further empower it with knowledge of real-world entities.
   We hope our study draws more attention to image ad understanding which is broadly relevant to the advertising industry. 
\end{abstract}

\section{Introduction} \label{sec:intro}
\begin{figure}
\centering
\includegraphics[width=\linewidth]{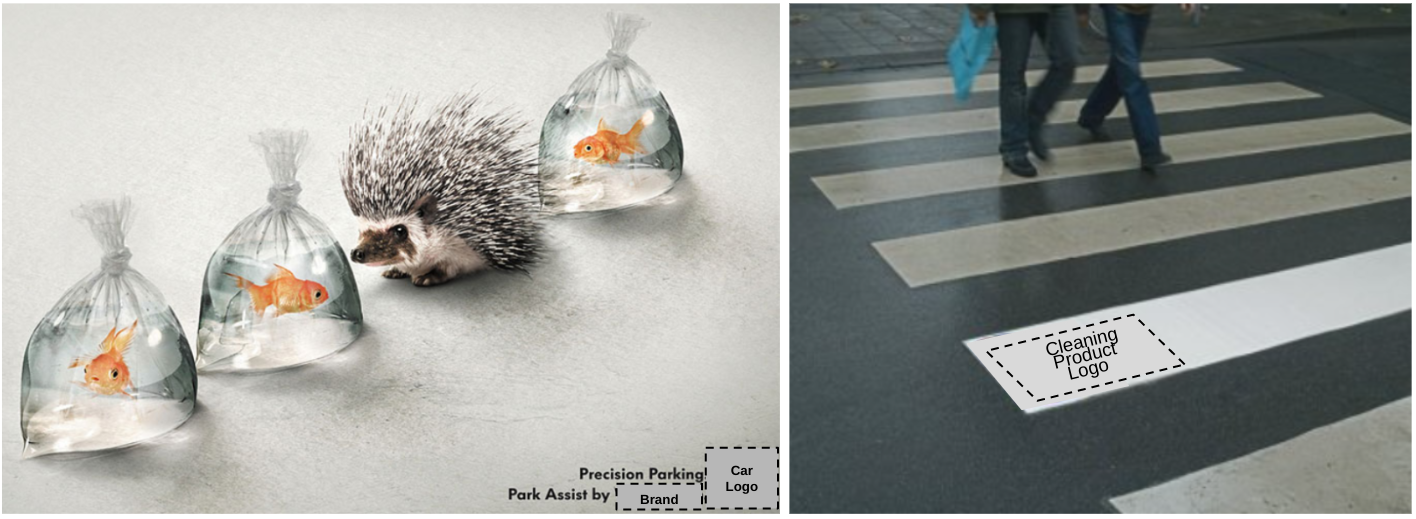} \\
\caption{Example image ads with diverse visual elements, atypical scenes and rhetorics to convey their messages creatively. All brand info is anonymized.}
\label{fig:ad_examples}
\end{figure}

As advertisements play an integral role in human society,
image ad understanding has many real-world applications such as ad targeting \cite{hussain2017automatic}, visual metaphor understanding \cite{abokhoza2019advertising} and creative ad generation \cite{chilton2019visiblends,akula2022metaclue}.
It is also highly challenging due to several reasons, as exemplified in Fig. \ref{fig:ad_examples}.
\textit{First}, image ads consist of diverse visual elements including non-photorealistic objects and atypical scenes synthesized creatively that are beyond common academic datasets.
\textit{Secondly}, they involve knowledge of a large number of real-world entities such as brands and products where existing work \cite{su2018scalable,li2022seetek} struggles to cover.
\textit{Lastly}, many adopt visual or multimodal rhetorics requiring reasoning over diverse visual elements including scene-texts, and sometimes even elude humans \cite{petridis2019human}.
However, image ad understanding is relatively under-explored in the machine learning community, especially in the presence of recently developed foundational vision-language models (VLMs) pre-trained using a tremendous number of image and text description data.

\begin{figure}
\centering
\includegraphics[width=0.98\linewidth]{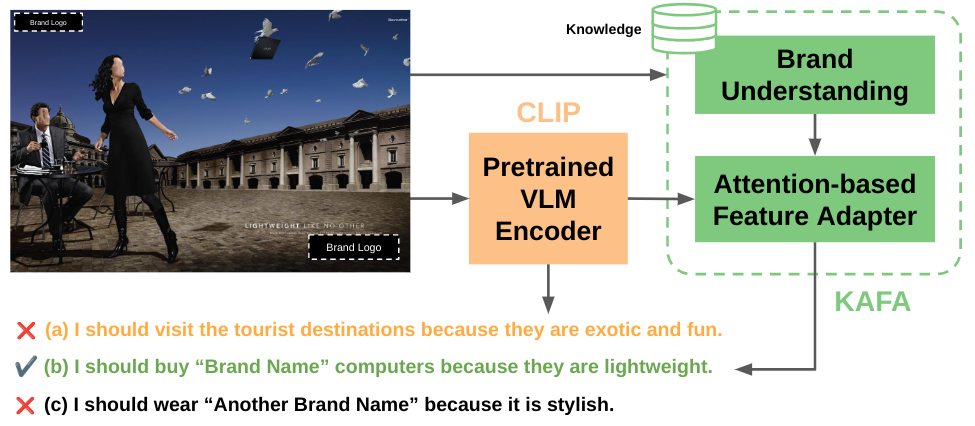} \\
\caption{We propose to utilize external knowledge via a brand understanding module and combine features of different modalities via a lightweight attention-based feature adapter to decode the correct messages of image ads. The VLM baseline is confused and gives the wrong one. All brand info is anonymized.}
\label{fig:teaser}
\end{figure}

The pre-trained VLMs are shown to have great generalization capability, contain real-world knowledge (implicitly), and can be adapted to a wide range of downstream tasks in a data-efficient way \cite{radford2021learning,alayrac2022flamingo}.
It is then natural to utilize VLMs for image ad understanding.
In this paper, we perform the first empirical study of adapting VLMs to the task of decoding the overall messages delivered by image ads, which is usually formulated as visual question answering \cite{hussain2017automatic}.
Specifically, we examine three popular pre-trained VLMs that are alignment-based and are publicly available, namely, CLIP \cite{radford2021learning}, ALBEF \cite{li2021align} and LiT \cite{zhai2022lit}.
We examine zero-shot performance as well as adaptation strategies and reveal the practical challenges of applying VLMs to image ads.
We propose a simple feature adaptation strategy that effectively utilizes VLM features.
We further propose to incorporate external brand knowledge (real-world entities) that brings a significant performance boost.

Our contributions are three-fold.
\textbf{First}, we empirically find that the sheer scale of data \& capacity of the model used in pretraining matters the most for the performance of image ad understanding, partly due to VLM's capability of storing real-world knowledge, which is not captured well by the commonly used metrics for comparing VLMs.
\textbf{Second}, we reveal the practical challenges of adapting VLMs for image ad understanding (i.e., overfitting to the limited training data \& supervision signals and high computation burden of hard negative mining) and propose a simple solution (attention-based feature adaptation) that better leverages VLM features than previous adaptation strategies.
\textbf{Lastly}, we propose to leverage external knowledge for brand understanding that we have empirically shown to further enhance image ad understanding.
Together with the aforementioned adaptation strategy, we call our approach knowledge-augmented feature adaptation (KAFA). 

\section{Related Work}

\paragraph{Image Ad Understanding} 
Learning to automatically interpret image ads was proposed by the Pitt Image Ads Dataset \cite{hussain2017automatic}, where each ad is annotated by a caption that answers ``what should I do according to the ad and why?''
Different from traditional image captioning, this task is highly non-trivial as discussed at the beginning of Sec. \ref{sec:intro}.
While prior methods utilize cultural connotations via external symbolic knowledge \cite{ye2018advise}, capture relations between scene-texts and objects by GNNs \cite{dey2021beyond}, and leverage pre-trained language models to combine multimodel information \cite{kalra2020understanding}, none have exploited vision-language models (VLMs) and the knowledge of real-world entities (i.e., brands).
Besides the wide applications in the ad industry, later work hints that the study of image ads is relevant to much broader research topics \cite{singh2019towards,akula2022metaclue}.

\paragraph{Foundational Alignment-based VLMs}
A recent surge of collections of tremendous images paired with text descriptions \cite{schuhmann2022laion} enables alignment-based pretraining (i.e., contrastive learning) of foundational VLMs that are efficient zero-shot or low-shot learners for downstream tasks.
By learning to embed images and texts into a shared semantic space, they handle domain variations in an open-vocabulary manner (which involves real-world knowledge).
Among these are CLIP \cite{radford2021learning}, ALIGN \cite{jia2021scaling}, LiT \cite{zhai2022lit} and BASIC \cite{pham2021combined}.
Another line of work further adopts masked language modeling, image captioning loss, and object-level alignment, e.g., ALBEF \cite{li2021align}, Florence \cite{yuan2021florence}, CoCa \cite{yu2022coca} and GLIP \cite{li2022grounded}.

\paragraph{Transfer Learning of VLMs}
Transfer learning of VLMs has become popular with the zero-shot performance of CLIP in  image classification tasks. 
A direct approach is to (partially) fine-tune the VLMs with (optionally) additional neural networks tailored for downstream tasks, e.g., TAP-C \cite{song2022clip}, CPT \cite{yao2021cpt}, KAT \cite{gui2021kat} and VL-Adapter \cite{sung2022vl}.
Another approach that bypasses the need of tuning the VLMs is prompt learning. 
For instance, CoOp \cite{zhou2022learning} and CoCoOp \cite{zhou2022conditional} only tune learnable inputs to the VLMs.
The third approach that further reduces memory and computation burden is feature adapters, where VLM features of the inputs are pre-computed before transfer learning.
Examples are CLIP-Adapter \cite{gao2021clip}, SVL-Adapter \cite{pantazis2022svl} and Attention-Adapter \cite{zhao2022tiny}.


\paragraph{Knowledge-Augmented Image Understanding}
Many image understanding tasks require real-world knowledge beyond what can be captured by the input data.
For instance, FVQA \cite{wang2017fvqa} and OK-VQA \cite{marino2019ok} require models to process external fact-based knowledge; TextVQA \cite{singh2019towards} asks to understand named entities in the wild; the Pitt Dataset \cite{hussain2017automatic} involves recognition of large quantities of brands.
Existing work incorporates external knowledge either explicitly via structured or unstructured knowledge base \cite{wang2015explicit,garderes2020conceptbert,ye2018advise}, or implicitly from knowledge stored in pretrained models \cite{kalra2020understanding,kim2022ask}, or both \cite{marino2021krisp,gui2021kat}.


\section{What Really Matters for Pre-trained VLMs in Image Ad Understanding?} \label{sec:what_matters}
The first insight of our empirical study is that the sheer size of data and the model used in pretraining is the key factor determining the performance of VLMs for image ad understanding. 

To promote reproducibility, we evaluate three alignment-based VLMs (i.e., CLIP, ALBEF and LiT) that are publicly accessible in a zero-shot manner on the Pitt Dataset \cite{hussain2017automatic}, which formulates ad understanding as image-to-text retrieval.
We adopt the official evaluation \href{https://eval.ai/web/challenges/challenge-page/86/evaluation}{protocol}, which asks the model to select one of the 3 correct messages conveyed by the image ad from a set of 15 candidates (including 12 wrong messages) for each of the 12805 test samples.
Specifically, given an alignment-based VLM, let us denote its encoders with normalized outputs as $f_{I}(\cdot)$ and $f_{T}(\cdot)$ for image and text branches, respectively.
Given an image $\textbf{x}$ and the ground truth texts $\textbf{y}$, the VLM retrieves $y$ from candidates $\mathcal{C}(\textbf{x})$ according to the dot-product score
$f_I(\textbf{x}) \cdot f_{T}(y)$.
We then measure the performance of the VLM with 3 metrics commonly used in the literature: \textit{accuracy} (the percentage of images with any positive
text retrieved with rank one), \textit{rank} (how the top retrieved positive text is ranked averaged over all images), and the \textit{mean rank} (the mean rank of the all positive texts averaged over all images).

With the results reported in Tab. \ref{tab:ads_official_numerical}, we have several findings.
\textit{First}, the more data used during the pretraining of a VLM, the better it generalizes to the image ad domain.
For a comparison, CLIP has seen 400M image-text pairs, LiT 100M, and ALBEF 14M.
\textit{Second}, the larger the capacity of a model, the better it understands image ads.
We have evaluated different sizes of the CLIP model beyond the three sizes shown in Tab. \ref{tab:ads_official_numerical} and the trend keeps the same.
\textit{Third}, commonly used metrics for comparing VLMs, including zero-shot accuracy on the ImageNet \cite{russakovsky2015imagenet} validation set (for which LiT claims to outperform CLIP) and image-to-text retrieval precision on Flickr30K \cite{young2014image} (for which ALBEF claims to outperform CLIP), do not reflect the performance of image ad understanding well.

We hypothesize that this is partly because image ad understanding requires knowledge of real-world entities (e.g., brands) which the pre-trained models contain.
Similar to the dramatic performance advancement of GPT language models \cite{brown2020language} driven by the larger scale of training data and the model capacity, more knowledge can be distilled and implicitly stored in the weights of pre-trained VLMs with larger models and more pre-training data.
We empirically verify that the VLM's capability of recognizing brands from images is aligned with its performance of decoding the messages from the ads. 
See results in Tab. \ref{tab:brand_results}.

\begin{table} 
		\centering
  \vspace{6px}
    \begin{adjustbox}{width=1.0\columnwidth}
  \begin{tabular}{lccc} 
    \toprule
     & Acc $\uparrow$ & Rank $\downarrow$ & m. Rank $\downarrow$ \\
    \midrule
    ViLBERT \cite{lu2019vilbert} & 61.8 & 1.860 & 4.190 \\
    VS (v1) \cite{dey2021beyond} & 86.8 & 1.264 & 3.072 \\
    BERT-FT \cite{kalra2020understanding} & 89.7 & 1.230 & 2.982 \\  
    \midrule
    ALBEF \cite{li2021align} & 57.6 & 2.220 & 4.935 \\
    ALBEF (ft. on Flickr30k) & 64.2 & 2.242 & 5.125 \\
    ALBEF (ft. on MSCOCO) & 64.0 & 2.002 & 4.651 \\
    LiT (L16L) \cite{zhai2022lit} & 64.0 & 1.849 & 4.268 \\
    CLIP (ViT-B/32) \cite{radford2021learning} & 88.1 & 1.213 & 2.937 \\
    CLIP (ViT-B/16) & 92.2 & 1.123 & 2.694 \\
    CLIP (ViT-L/14@336px) & 95.2 & 1.069 & 2.547 \\
    \midrule
    KAFA (ours) & \textbf{97.4} & \textbf{1.033} & \textbf{2.391} \\
    \bottomrule
  \end{tabular}
  \end{adjustbox}
	\caption{Zero-shot VLM performance on the Pitt Dataset \cite{hussain2017automatic} with its official eval protocol (3 positive texts and 12 negative ones for each test image). The best CLIP model already surpasses previous state-of-the-art results (BERT-FT). The size of the data and model used in VLM pretraining have a huge impact on the results. See Sec. \ref{sec:what_matters} for details of the metrics. For completeness, we also include the results of our proposed method (KAFA) here.}
	\label{tab:ads_official_numerical}
\end{table}

\section{Challenges in VLM Adaptations to Image Ads and An Intuitive Solution} \label{sec:adaptation}
With CLIP as the clear champion, we further study VLM adaptations for image ad understanding using the best CLIP model (ViT-L/14@336px) as the backbone. 
We aim to enable better performance for image ad understanding by better adapting pre-trained VLMs to the image ad domain with the help of additional information such as scene-texts extracted from the image.

\subsection{The Issue of Overfitting and High Computation Complexity}
We find two practical challenges in adapting pre-trained VLMs to the image ads, \textit{first}, the overfitting issue in fine-tuning due to limited image ads and the lack of a strong supervision signal, and \textit{second}, the high computation burden caused by solutions to the previous challenge.

Annotations of image ads are hard to obtain in general \cite{akula2022metaclue}, making it common to only have limited training data (e.g., the Pitt Dataset only contains 51,223 image-text pairs).
This results in VLM's vulnerability to overfitting during adaptation.
We find that directly fine-tuning CLIP contrastively on the Pitt Dataset with the symmetric cross-entropy loss (as in the original CLIP paper) gives worse performance than the zero-shot one unless we adopt early stopping and a carefully tuned learning rate schedule.
Moreover, as reported in Tab. \ref{tab:ads_official_numerical}, the best zero-shot performance of CLIP already surpasses the previous state-of-the-art and is very close to $100\%$, leading to very weak supervision signals for vanilla fine-tuning.
We thus need strong training signals.
To save GPU memory required by much larger batch sizes, we adopt hard negative mining \cite{xuan2020hard}, which selects hard negatives from a very large candidate set as opposed to within the mini-batch.

However, hard negative mining (HNM) strategies usually incur a large computation burden.
In fully online hard negative mining (denoted full HNM), for each training image $\textbf{x}$ and the corresponding texts $\textbf{y}$, we first rank $N_{cand}$ negative texts $\{y | y \neq \textbf{y}\}$ sampled from the entire training data according to the online similarity scores (the dot-product score
$f_I(\textbf{x}) \cdot f_{T}(y)$ computed from the current VLM model), and then we choose the $N_{hard}-1$ most similar $y$ as the hard negatives.
While this essentially constructs a much harder candidate set $\mathcal{C}(\textbf{x})$, it requires the computation of features of all training texts at every gradient step, which is prohibitively expensive.
Existing methods propose to reduce the complexity by keeping a sparsely updated bank of all training text features (memory bank) \cite{wu2018unsupervised} or with the help of a momentum-updated text encoder (MoCo) \cite{he2020momentum}.
Nevertheless, we tailor these methods to our setup\footnote{We use these methods to compute similarity scores but still only select the hardest negatives for fine-tuning to save GPU memory (the purpose of HNM).} and find that they perform worse than full HNM.
We report the accuracy (\%) in Tab. \ref{tab:hnm} with a harder eval protocol than the official one by using larger numbers ($K$) of negative samples randomly drawn from the test texts (thus a set of harder negatives).

\begin{table}[ht]
  \centering    
  \footnotesize
      \begin{adjustbox}{width=0.96\columnwidth}
 \begin{tabular}{lcccc}
    \toprule
     Number of Candidates $K$ & 20 & 100 & 500 & 1000 \\
    \midrule
    Zero-shot & 91.7 & 80.7 & 64.4 & 56.5 \\
    Direct FT & 92.4 & 82.2 & 66.7 & 59.0 \\
    Direct FT + memory bank & 92.8 & 82.9 & 67.5 & 60.3 \\
    Direct FT + MoCo & 93.3 & 83.8 & 69.8 & 62.4 \\
    Direct FT + full HNM & 93.7 & 84.6 & 70.0 & 62.9 \\
    \bottomrule
  \end{tabular}
  \end{adjustbox}
\caption{Accuracy (\%) reported with different sizes ($K$) of the candidate set on the test set of the Pitt Dataset. The larger $K$ means harder negative samples. Zero-shot is the zero-shot performance of the best CLIP model. FT means fine-tuning the best CLIP model.}
\label{tab:hnm}
\end{table}


We believe this is because image ad understanding requires fine-grained information extraction (e.g., the specific brand of a product) and both these two strategies are subject to the destruction of such information as they compute the loss not in a fully online manner.
In particular, their text features used for contrastive fine-tuning always come from different VLM encoders, either the past checkpoints or the momentum-updated versions). 
Although direct fine-tuning with full HNM outperforms the others, it is extremely inefficient and thus impractical.

\begin{figure*}
\centering
\includegraphics[width=\linewidth]{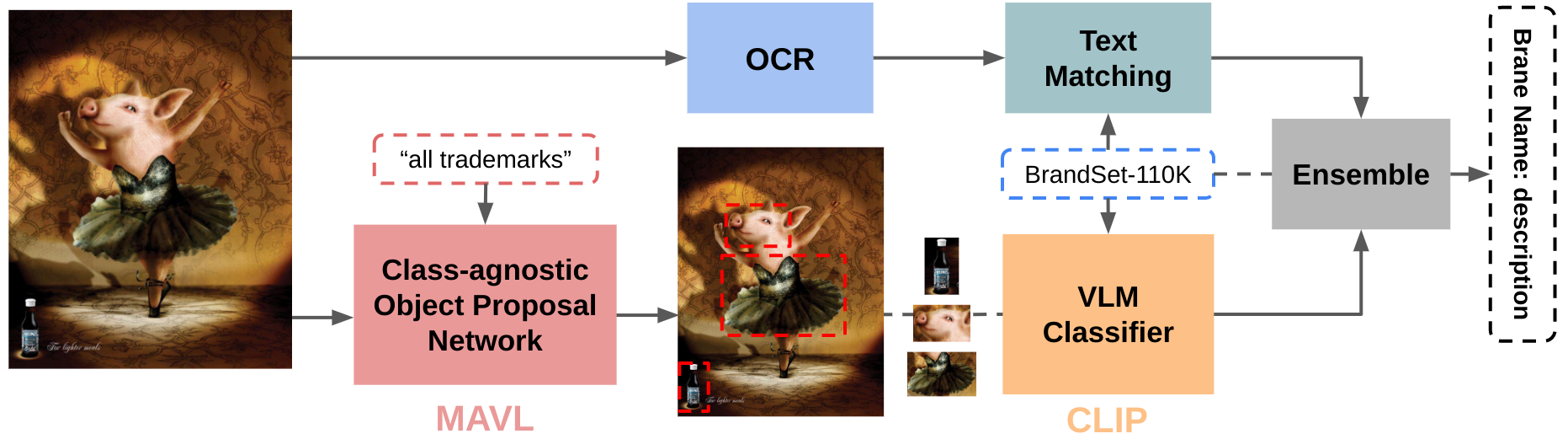} \\
\caption{Illustration of our brand understanding module that is an ensemble of text-matching and vision-based recognition. Given an input image ad, we use MAVL to propose regions by prompting ``all trademarks'' and retrieve entries in BrandSet-110K over the regions with CLIP.
We aggregate the predictions across regions and the text-matching results to generate the final output via some simple rules (see details in Appendix \ref{sec:brand_recognition}).}
\label{fig:knowledge_retriever}
\end{figure*}

\subsection{Feature Adaptation as the Solution} \label{sec:feature_adaptation}
We propose a simple and intuitive solution, attention-based feature adaptors, that both handle the aforementioned issues during adaptations and enable incorporating additional information (e.g., scene-texts) for better image ad understanding. 

Feature adapters are recently proposed \cite{gao2021clip, zhang2021tip, pantazis2022svl,zhao2022tiny} as a line of very efficient adaptation strategies of VLMs.
They freeze the weights of the pretrained VLMs, pre-compute features using their encoders, and use additional lightweight adapter networks to process these features.
As a result, on-the-fly feature computation over a massive candidate set becomes computationally feasible and so is the fully online hard negative mining, since we only compute the adapted features online via a lightweight network.
More efficiently, we can set the text adapter to an identity function (i.e., only use adapters for image features).

More importantly, feature adapters are suitable for fusing info from multiple sources.
While previous feature adapters are mostly designed for image classification, we consider it as a strategy to aggregate multiple input branches (of potentially different modalities).
For instance, previous methods for image ad understanding, such as VS \cite{dey2021beyond}, utilize scene-texts extracted from images (by OCR) to enhance its performance.
Similarly, we can extract text features from scene-texts using a VLM's text encoder and merge them with the image features extracted by the image encoder (of the same VLM) via a feature adapter.
In doing so, we obtain a better representation of image ads.

\begin{figure*}
\centering
\includegraphics[width=0.95\linewidth]{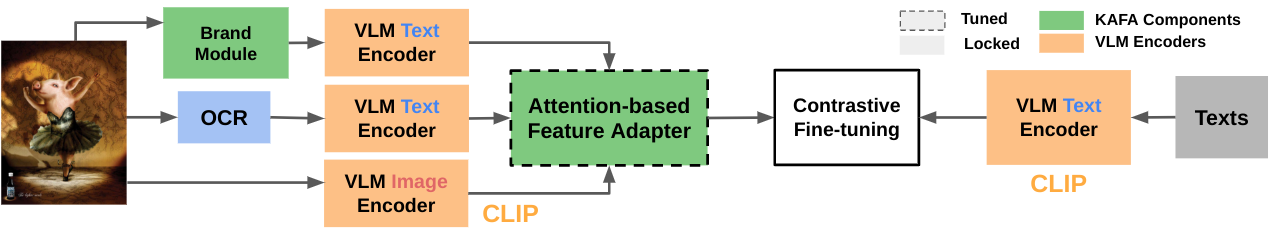} \\
\caption{The overall training pipeline of our proposed KAFA, where three branches of information are fed into the attention-based feature adapter, the only neural module free in the fine-tuning process. We leverage VLM encoders for both sides of the contrastive fine-tuning.}
\label{fig:adaptation_pipeline}
\end{figure*}

Specifically, we propose to adopt one layer of multi-head attention \cite{vaswani2017attention} as our feature adapter design, similar to the Tiny-Attention Adapter \cite{zhao2022tiny}.
Here the input sequence to the attention layer varies by modalities (brand, scene-texts and image, as in Fig. \ref{fig:adaptation_pipeline}) instead of temporally or spatially as commonly in Transformers.
By the nature of alignment-based VLMs, all information (whether in the text format as the scene-texts or the visual elements) are embedded as vectors and lie in a shared semantic space. 
We then utilize this property and fuse complementary information (e.g., image features and scene-text features) into one feature.
Moreover, we append a linear layer after the attention features and equip it with a residual connection.
Let us use the notation in previous sections and further denote $x_{st}$ as the scene-texts extracted from the image $\textbf{x}$ (by Google OCR \href{https://cloud.google.com/vision/docs/ocr}{APIs}).
Then our adapter is represented as  
\begin{equation*}
f^{att}(\textbf{x}) = \texttt{n}(f_I(\textbf{x}) + \mathcal{A}[f_I(\textbf{x}), f_T(x_{st}), ...][0])
\end{equation*}
where $\texttt{n}(\cdot)$ is a normalization function and $\mathcal{A}$ is multi-head attention (we leave room for other input branches by leaving ``...'' here).
Note that we do not use any adapter for the text descriptions of images (the labels of image ads), which further reduces the computation complexity as now we only need to compute and cache all text features in the training set once and for all during full HNM. 

In comparison, we also evaluate the popular CLIP-Adapter \cite{gao2021clip} as a strong baseline, which we tailor to our setup by training three 2-layer residual MLPs.
Please see the Appendix for implementation details.
As reported in Tab. \ref{tab:main_results}, our proposal of using an attention-based adapter (denoted KAFA w/o K) utilizes VLM features well by aligning multimodal features already in the same semantic space and outperforms CLIP-Adapter.
While other existing work \cite{shen2021much,gui2021kat} merges multiple branches of information by leveraging foundation models, they rely on large encoder-decoder networks that are computationally intensive and might not work well with limited training data as in our case.


\section{Improving Image Ad Understanding with External Knowledge}
To further improve image ad understanding, we propose to leverage external knowledge of real-world entities, namely product and brand information.
The major focus of advertisements is to promote brand awareness \cite{macdonald2003management}.
Sometimes brand information is even a necessity to interpret ads correctly since it eliminates ambiguities and gives visual cues to the audiences (e.g., the ad for a cleaning product in Fig. \ref{fig:ad_examples}).
It is then natural to empower feature adapters introduced previously with a brand understanding module that extracts brand information from images.
Here we present our training-free brand understanding module that considerably exploits VLMs.

\subsection{Brand Understanding Module} \label{sec:knowledge_retriever}
Extracting brand information from an image is very challenging due to 
the sheer scale of brands in the real world.
Existing published work \cite{su2018scalable,li2022seetek} and even commercial APIs tend to fall short of a good coverage.
To solve this issue, we construct a knowledge base that covers brands much better than existing datasets.
Our knowledge base has the format, \textit{KFC}: \textit{KFC is a fast food chain}, with around 110k entries covering names of brands, companies, organizations and others appearing in image ads.
We call this dataset BrandSet-110K (see details in Appendix \ref{sec:brand_recognition}).

Next, we take an ensemble approach to detect and retrieve relevant brand entries from BrandSet-110K given an image ad.
On one hand, we retrieve brands by performing string matching over all names in BandSet-110K using the scene-texts extracted by OCR from the image.
On the other hand, in case of OCR failures, no detection (some logos have no texts), or multiple detected entries (potentially false positives as most image ads promote only one brand at a time), we use a more powerful vision-based module.
Specifically, we adopt MAVL \cite{maaz2022class}, a state-of-the-art VLM, to propose object regions according to the text prompt ``all trademarks'',
We then use the best CLIP model to perform region classification based on a set of carefully engineered prompts.
And then, we select the best entries in BrandSet-110K according to the proposed regions.
We finally use some simple rules to combine the retrieved results from text-matching and the vision-based module, as in Fig. \ref{fig:knowledge_retriever} (see details in the Appendix).

Overall, our brand understanding module is training-free, covers much more entities than previously published work, and even outperforms some commercial logo detection APIs by evaluation on a small validation set, as reported in Tab. \ref{tab:brand_results}

\begin{table}[ht]
  \centering    
  \footnotesize
      \begin{adjustbox}{width=0.96\columnwidth}
 \begin{tabular}{lccccc}
    \toprule
     Method & Inputs & 20 & 100 & 500 & 1000 \\
    \midrule
    Zero-shot & I & 91.7 & 80.7 & 64.4 & 56.5 \\
    Direct FT + full HNM & I & 93.7 & 84.6 & 70.0 & 62.9 \\
    CLIP-Adapter & I+ST & 93.9 & 85.0 & 70.2 & 62.8 \\
    \midrule
    KAFA w/o K & I+ST & 95.0 & 86.8 & 72.7 & 65.1 \\
    KAFA w/o ST & I+K & 94.7 & 86.5 & 72.3 & 64.5 \\
    KAFA (ours) & I+ST+K & \textbf{95.6} & \textbf{87.7} & \textbf{73.9} & \textbf{66.0} \\
    \bottomrule
  \end{tabular}
  \end{adjustbox}
\caption{Accuracy (\%) reported on the Pitt Dataset. KAFA (our proposed attention-based adapter with external knowledge) achieves the best results compared to other approaches and the versions with fewer inputs (K = brand knowledge, ST = scene-texts, I = image). Note: ``Direct FT + full HN'' is extremely inefficient.} \label{tab:main_results}
\end{table}

\begin{table}[ht]
  \centering    
  \footnotesize
      \begin{adjustbox}{width=0.95\columnwidth}
    \begin{tabular}{lclc}
    \toprule
     &  Acc (\%) & & Acc (\%)\\
    \midrule
    VLM-based (ALBEF) & 14.5 & Text-matching & 36.0 \\
    VLM-based (LiT) & 29.0 & Google Cloud API &  42.0 \\
    VLM-based (CLIP) & 64.4 & Combined (Text + CLIP) & \textbf{66.6} \\
    \bottomrule
  \end{tabular}
  \end{adjustbox}
  \caption{Brand recognition accuracy on \textasciitilde 600 validation image ads. It justifies our brand understanding module and further verifies that models better at recognizing brands are better at image ad understanding.} \label{tab:brand_results}
  \vspace{-5pt}
\end{table}

\subsection{Overall Pipeline and Final Results}
Combining with our proposed brand understanding module, we illustrate our overall pipeline in Fig. \ref{fig:adaptation_pipeline} and call this approach knowledge-augmented feature adaptation (KAFA). 
In Tab. \ref{tab:main_results}, we demonstrate that KAFA achieves substantial improvements in image ad understanding over the VLM baseline and consistently outperforms other ablation versions with fewer inputs, justifying that our proposed brand understanding module helps to further improve image ad understanding.
We present an example in Fig. \ref{fig:teaser} to illustrate the improvement of our method over the baseline, where for better display we only show 2 negative text descriptions.
See more examples in Appendix \ref{sec:more_examples}.

\section{Additional Analysis}

\subsection{Hard Negative Samples in Evaluations} 
\begin{figure}
\centering
\includegraphics[width=0.95\linewidth]{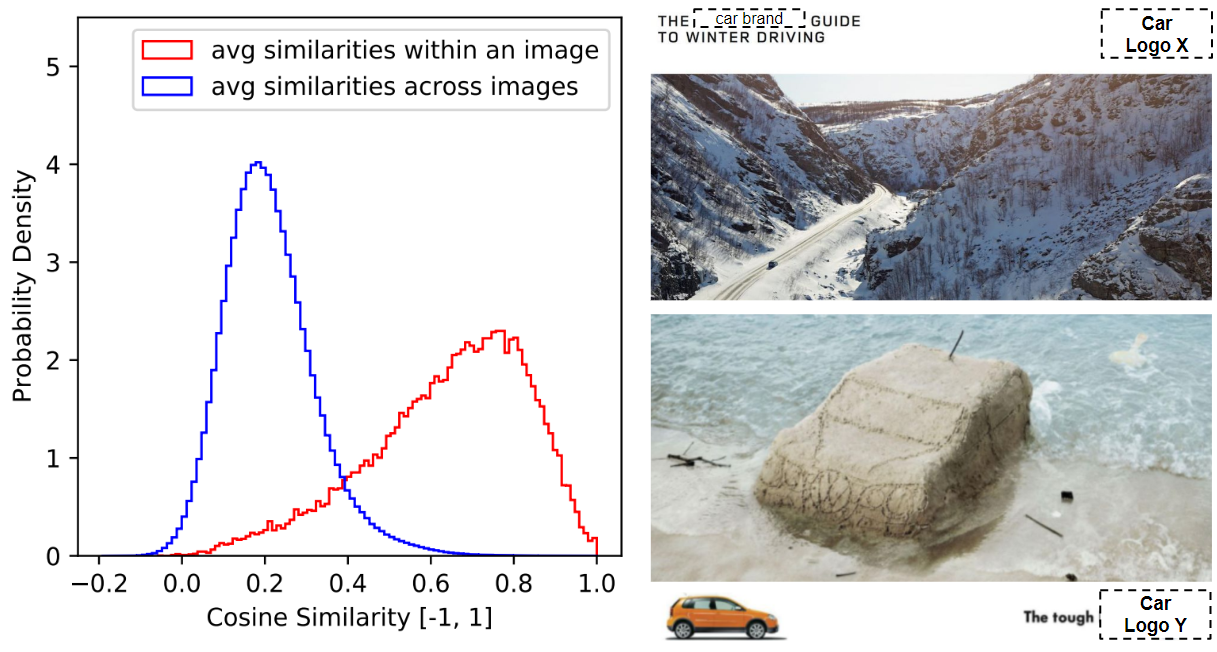} \\
\vspace{-3pt}
\caption{(\textbf{Left}) Similarity distributions of texts of the same and across images. Both are spread out with no easy cutoff threshold to sample hard negatives.
(\textbf{Right}) Two different ads share the same message ``I should drive this car because it can drive anywhere'', exemplifying the difficulty of sampling hard negative samples.}
\label{fig:collision}
\vspace{-7pt}
\end{figure}

We report our main results with a harder eval protocol than the official one. 
In fact, it is a challenge to perform effective evaluations in retrieval tasks \cite{akula2020words}.
While we need \textbf{hard} negatives to better reflect the capabilities of a model, usually by increasing the size of the candidate set, we also want those hard negatives to be real \textbf{negatives}. 
As illustrated in Fig. \ref{fig:collision} (right), two companies can have two different image ads that share a very similar message.
Hence, given an image, simply using a text of another as the negative might not work.

There is no easy solution.
We can use a generic sentence encoder to measure similarities among different texts in \cite{hussain2017automatic} and only sample texts that are semantically different from the target one (the ground truth) as negatives.
We adopt a strong sentence encoder (publicly available \href{https://huggingface.co/sentence-transformers/all-MiniLM-L6-v2}{here}) based on MiniLM \cite{wang2020minilm} to measure semantic similarities.
We compute similarities among descriptions of the same ad and those across different ads.
The similarity distributions are spread out, as demonstrated in Fig. \ref{fig:collision} (left), without easy cutoff thresholds to make negative samples both hard and truly negative.
Instead, we propose to use several different sizes $K$ of the candidate set with $K=20,100,500,1000$.
For each image in the Pitt Dataset \cite{hussain2017automatic}, we randomly choose a text from the ground truth and uniformly sample $K-1$ negatives from other images (harder negatives with larger $K$).

While most existing methods evaluate \cite{hussain2017automatic} with the official evaluation \href{https://eval.ai/web/challenges/challenge-page/86/overview}{protocol} (for ease of comparison we also provide results by this protocol in Tab. \ref{tab:ads_official_numerical}), it suffers from the lack of hard negatives.
Each image ad comes with only 15 randomly sampled candidate texts including 3 positives, giving a random model a 20\% accuracy.
Moreover, negatives are easy as they tend to be semantically distinct from the positives, making it hard to examine a model at finer levels.
We provide examples to compare negatives sampled in our protocol and in the official one in Appendix \ref{sec:negative_samples}.

\subsection{Data Leakage Regarding VLMs}
\begin{figure}
\centering
\includegraphics[width=\linewidth]{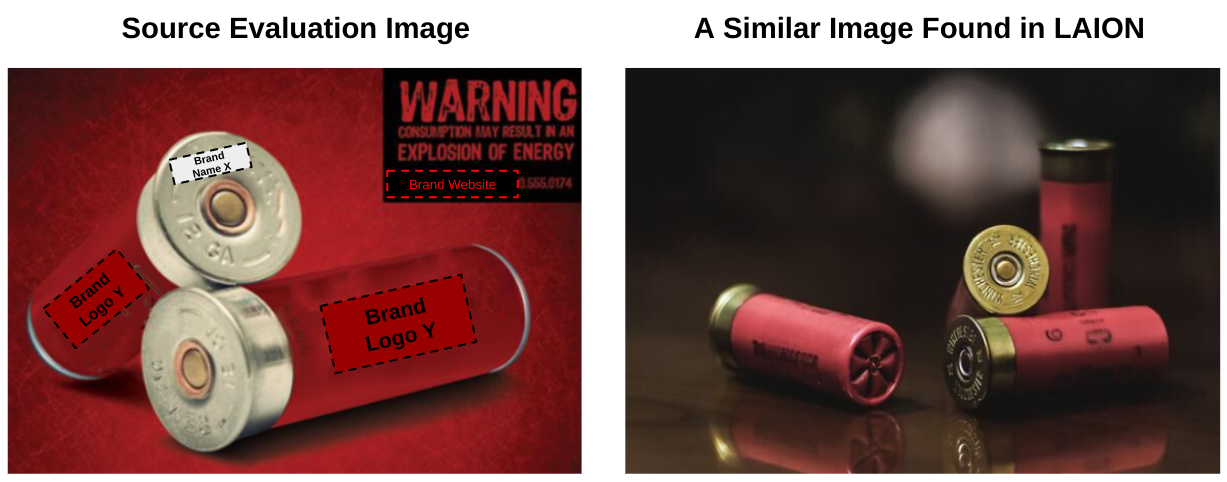} \\
\caption{An evaluation image and a found one in LAION-400M. As a reference, this image's caption reads: I should drink ``Brand Name'' because it'll give me a recharge of energy.}
\label{fig:data_leakage}
\end{figure}

The CLIP \cite{radford2021learning} model we use in our experiments was pre-trained on a tremendous amount (400M) of image-text pairs on the Internet.
A concern is that there might be data leakage, i.e., the pre-trained VLMs might have already seen images in the evaluation set, leading to inflated results.
We perform an analysis to conclude that this is unlikely the case.
We manually inspect images in the LAION-400M dataset \cite{schuhmann2021laion} that are semantically similar to a set of randomly sampled 100 eval image-text pairs.
While the dataset used to train CLIP is not publicly released, LAION-400M is a very close one with a similar scale of data filtered by the CLIP model.
Specifically, for each of the 100 random samples, we use the open-sourced CLIP-retrieval tool (\href{https://github.com/rom1504/clip-retrieval}{here}) to find the closest images from LAION-400M indexed by both the sample text and image.
We do not find any substantially overlapped content or near duplicates (see Fig. \ref{fig:data_leakage} as an example).
Moreover, our proposed method achieves significant performance improvement over the VLM baseline and both are based on the same CLIP model.
Therefore, data leakage is less of a concern.

\section{Conclusion}
In this paper, we study the adaptation of pretrained alignment-based VLMs for the challenging image ad understanding task.
We benchmark and reveal practical challenges in adapting VLMs, propose a simple and intuitive (yet effective) strategy for feature adaptations, and further improve image ad understanding with external brand knowledge.
While we mainly focus on the image-to-text retrieval task for its simplicity, we believe further studies can extend it to directly generating text descriptions given image ads or even generating image ads given the descriptions.
We hope our study draws more attention to image ad understanding that are relevant to the advertising industry and provide insights for a broader machine learning community. 
   
\section*{Limitations}
The data from the Pitt Dataset \cite{hussain2017automatic}, while useful for our paper, contains many images and annotations that may perpetuate harmful stereotypes according to sensitive characteristics such as gender and carry the risk of amplification by machine learning models. We plan to collaborate with AI robustness researchers to identify such examples and develop methods for improving ML models in terms of robustness and reliability.

\bibliography{anthology,custom}
\bibliographystyle{acl_natbib}

\newpage
\appendix

\section{Scene-text Extraction by OCR}
In our paper, we use scene-texts as one of the inputs for experiments in Pitt Dataset \cite{hussain2017automatic}.
We use the Google Cloud OCR API (\href{https://cloud.google.com/vision/docs/ocr}{link}) to extract all text tokens, which are grouped by paragraphs by the API.
We then group paragraphs into blocks by simple heuristic rules (e.g., two consecutive paragraphs with similar font sizes should be considered in the same block) and then filter out those blocks with an average prediction confidence score (provided by the API) less than $0.7$.

\section{Brand Recognition} \label{sec:brand_recognition}
\subsection{BrandSet-110K}
We construct BrandSet-110K by first compiling entries from public websites.
Specifically, for the list of topics (such as automobiles and healthcare) in the Pitt Dataset \cite{hussain2017automatic}, we Google with the query ``Top XX brands/companies in Y'' to obtain a list of thousands of common brands, organizations, etc., denote source I.
We further scrape the Google Knowledge Graph Search API (\href{https://developers.google.com/knowledge-graph}{link}) to find a much larger list of named entities, denoted source II, whose categories fall into ``brands'', ``companies'', etc., where each entry comes with a one-paragraph description.
Since results from the Knowledge Graph (KG) is a little bit noisy and might miss some popular entities, we rely on source I to make sure that the most prevalent entities appearing in our commercial world are included in our dataset.
We then query entries from source I in KG to also obtain the descriptions.
If such entries are not found in KG, we simply use the descriptions ``X is a brand name in the industry of Y''.
Together with source II, we obtain a raw combined knowledge base.
Then we filter out those entries that are common English words (if the entry appears in the English dictionary (\href{https://pypi.org/project/PyDictionary/}{link}) or a word set from NLTK \href{https://www.nltk.org/_modules/nltk/corpus.html}{(link)}).
We do so to remove entries such as ``Everyday'', which will result in too many false positives during brand detection.
We also remove entries consisting of a single character.
Eventually, we end up with around 110K entries, i.e., name and description pairs.

Since the descriptions returned by KG can be quite long, we further use a learning-based sentence parser to only select the very first sentence of the description (usually in the format of ``X is a brand/company/org in the industry of Y with Z features'').
We use this API (\href{https://huggingface.co/spacy/en_core_web_md?text=My+name+is+Clara+and+I+live+in+Berkeley%2C+California.}{link}) from Hugging Face \cite{wolf2019huggingface}, which is based on \href{https://spacy.io/}{spaCy}.

\subsection{Brand Recognition by Text-Matching}
The text-based brand recognition module essentially performs text matching to exhaustively search over all entries in BrandSet-110K given the scene-texts extracted by OCR.
For each name in BrandSet-110K that is larger than 6 characters, we match the text in a case-insensitive manner; otherwise, we match it case-sensitively to reduce false positives.
A name is set to be matched in a scene-text if it is a phrase of the text (``abc'' is matched in ``abc def'' but not in ``abcdef''.) 
When doing ablation studies of evaluating text-matching only performance, in case of multiple predictions we randomly select one as the output.

\subsection{Vision-based Brand Recognition}
The vision-based brand recognition module handles situations where the text-based one fails (when texts are too small or blurred or artistic for OCR to work; or when logos are purely graphic).
The vision-based module is a pipeline of several steps.
The class-agnostic region proposal (we use the best model in MAVL \cite{maaz2022class}, a state-of-the-art model) is adopted to generate candidate regions that contain brand logos or visual elements revealing brand information.
We choose ``all trademarks'' as the best prompt with other candidates such as:
\begin{itemize}
    \item ``all small objects'', ``all brand logos'',
    \item ``all brand icons'', ``all brands'', ``all logos''
\end{itemize}
After the region proposal, we use the best CLIP \cite{radford2021learning} model (its visual encoder) to compute the region features.
We include the entire image as an extra proposed region.
Then we use the text features (via the CLIP text encoder) of the following 6 prompts to find the best entry in BrandSet-110K. Namely
\begin{itemize}
    \item ``A brand logo of X'', ``A logo of X'',
    \item ``A trademark of X'',``A brand logo of X. Y'',
    \item ``A logo of X. Y'', ``A trademark of X. Y''
\end{itemize}
where X is the name and Y is the corresponding description in BrandSet-110K.
We first average dot products of the region features and brand features across all 6 prompts.
We then find two candidates: (1) the name X with the largest predicted scores among all names and all regions of an image and (2) the name X with the largest predicted scores averaged across all regions among all names that are champions in at least one region.
Our final output is chosen by the higher value of the dot products of the global image feature and the two text features of the prompt ``'An advertisement of X'' (we select this prompt after another minor prompt engineering process).  

\subsection{Ensemble of Text-matching and Vision-based Brand Recognition}
We use simple heuristic rules to ensemble the text-matching results and the vision-based ones.
Specifically, if there is no name detected from text-matching, we return the vision-based result; if there is only one name detected from text-matching, we return the text-based result; if more than one name is detected from text-matching, we select the name from detection of both text and vision-based modules by the highest value of the dot product of the global image feature and the text features of ``'An advertisement of X''.
The ensemble module finally returns the single name and the corresponding description in BrandSet-110K.

\section{Network Architecture of Attention-based Feature Adapter}
We adopt a very lightweight network for feature adaptation.
For each modality of the inputs (e.g., inputs to KAFA in the Pitt Dataset are three vectors: scene-text features, image features, and brand features), we first add learnable positional embedding (which is used to distinguish between different modalities) and then apply a multi-head attention layer \cite{vaswani2017attention} to obtain a list of vectors; we finally use the first vector (corresponding to the image feature input branch) and add residual connections from the input image feature (before positional embedding) to produce the final output (with normalization).
To make things clearer, we also provide the pseudocode.
\begin{lstlisting}[language=Python,
  % numbers=left,
  % stepnumber=1,
  numbersep=19pt,
  tabsize=2,
  showspaces=false,
  showstringspaces=false,
  basicstyle=\footnotesize
]
import torch.nn.Parameter as param
import torch.nn.functional as F

# args is a list of input features
# e.g., [img_fs,scene_text_fs,brand_fs]

pos_emb_list = []
for _ in range(n_input):
    pos_emb_list.append(
        param(torch.zeros([input_d])))
attn = torch.nn.MultiheadAttention(
    embed_dim=input_d, 
    num_heads=8, 
    batch_first=True)

inputs = []
for i in range(n_input):
    inputs.append(
        args[i] + pos_emb_list[i])
x = torch.stack(inputs, 1)
x, _ = attn(x, x, x, need_weights=False)
# The first is the image features.
x = x[:, 0] + args[0] 
x = F.normalize(x, dim=-1)
\end{lstlisting}

\section{Data Cleaning on Pitt Dataset}
We perform data cleaning on both the training and evaluation data of the Pitt Dataset (when evaluated using the official evaluation protocol, whose issue is discussed in the main paper, we stick to the raw evaluation set).
For every text in the dataset (the response to the ``what should I do according to the ad and why'' question), we remove invalid ones (e.g., ``I don't know'', ``not an ad'', ``not sure''), fix typos (e.g., ``becasue'', ``becaues''), and remove those without answering the ``why'' question.
Furthermore, we filter out texts that do not mention nouns or only have nouns that are not very informative (we compile a list of non-informative nouns appearing frequently in the dataset, such as ``product'', ``thing'' and ``vendor'').
This step is to remove non-specific texts such as ``I should buy this product because ...''.
In the end, we randomly select one text (with a fixed random seed) as the ground truth of its image.
If an image has all its texts removed by data cleaning, we remove the image from the dataset.
We find such images constituting less than 3\% of all images.

\section{Hard vs. Easy Negatives for Evaluation in Pitt Dataset} \label{sec:negative_samples}
\begin{figure}
\centering
\includegraphics[width=0.7\linewidth]{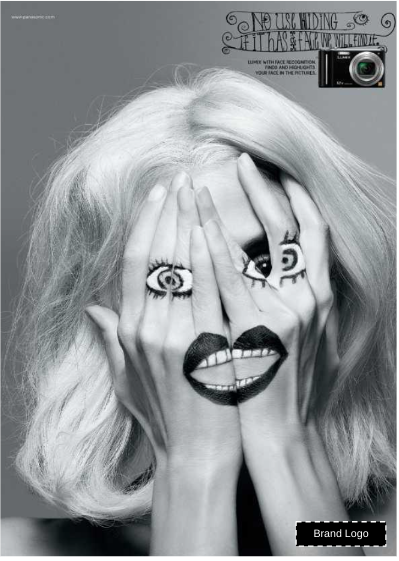} \\
\caption{An example to illustrate the issue of easy negative samples in evaluation.}
\label{fig:exapmle_in_appendix}
\end{figure}

Here we explain why we use larger number of candidates $K$ during evaluation.
Model evaluation for cross-modal retrieval is challenging \cite{akula2020words}.
The official evaluation protocol in Pitt Dataset suffers from the issue that it lacks hard negatives to fully reflect the perception and reasoning capability of the models.
Each image in the protocol has 3 positive texts and only 12 negative ones, giving a random guess model a 20\% accuracy.
On the contrary, increasing the number of candidates in our evaluation protocol as introduced in the main paper effectively yields harder negatives.
For instance, for the image ad in Fig. \ref{fig:exapmle_in_appendix} whose ground truth is ``I should buy a Brand A camera because it will help me create'', if we set the number of candidates to be 10 (i.e., 9 negatives), the best CLIP model makes the correct selection with all easy negatives, among which the most confusing ones are 
\begin{itemize}
    \item ``I should drink Brand B because it  de-ages you''
    \item ``I should not drown in my decision because bad choices will keep you under''
    \item ``I should buy this bag because it is resealable''
\end{itemize}
If we set 50 total candidates (i.e., 49 negatives), again the CLIP baseline predicts correctly with the most confusing ones still being relatively easy negatives:
\begin{itemize}
    \item ``I should use Brand C cosmetics because it makes you beautiful''
    \item ``I should buy Brand D products because they are reliable''
    \item ``I should see a movie because it's fun''
\end{itemize}
For a larger number (e.g., 100 total candidates), the CLIP model starts to make mistakes, with hard negatives such as 
\begin{itemize}
    \item ``I should use Brand E makeup because it will make me more seductive''
    \item ``I should buy Brand F makeup because it will make me beautiful''
    \item ``I should buy this makeup because it will make me shine''
\end{itemize}
Notice that for privacy reasons, all brand names in this example are anonymized.

\begin{figure*}
\centering
\includegraphics[width=\linewidth]{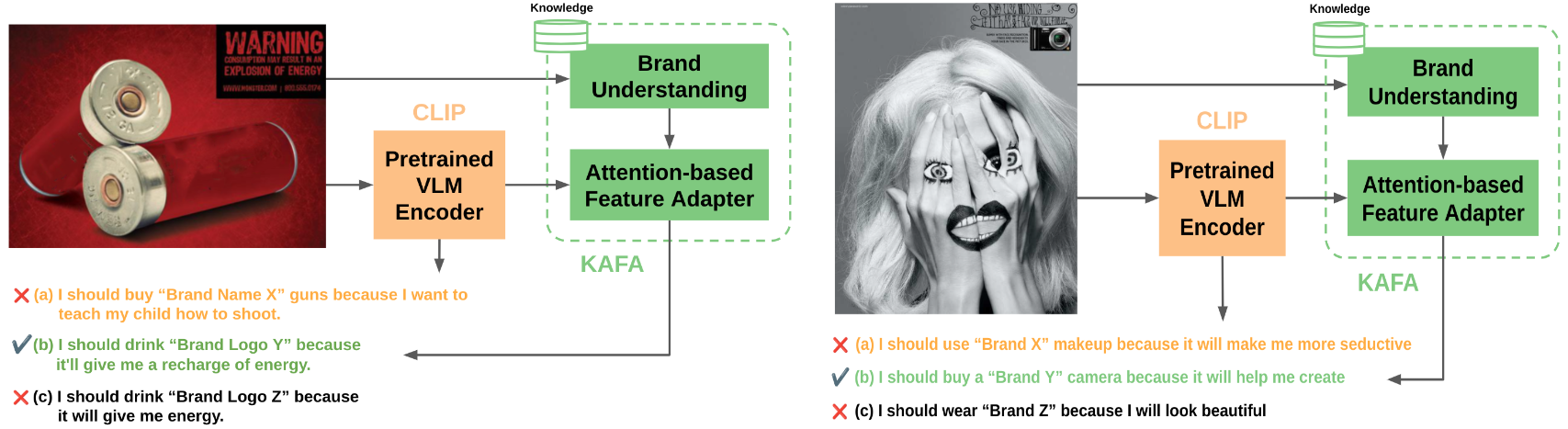} \\
\caption{Additional examples that demonstrate KAFA's improvements over the VLM baseline.}
\label{fig:more_examples}
\end{figure*}

\section{Training Details}
\subsection{Direct Fine-tuning of CLIP}
We fine-tune the best CLIP model on the training images of Pitt Dataset with a batch size of 8, symmetric cross-entropy loss (the one used in the original paper of CLIP) and the Adam optimizer \cite{kingma2014adam} with weight decay of $1e-4$. 
We set other parameters of Adam as in the original implementation of CLIP.
We find that using a very small learning rate (e.g., $1e-7$) is necessary for fine-tuning CLIP on Pitt Dataset; otherwise, the CLIP model can overfit easily.
For the same reason, we adopt early stopping and only fine-tune the model for a maximum of 4 epochs.
We leave the details in the next section for the fine-tuning version with online hard negative mining (very computationally intensive as suggested in the main paper).

\subsection{Fully Online Hard Negative Mining (full HNM)}
When performing hard negative mining during training, for each image in a mini-batch, we first compute the VLM features of a large number of randomly sampled negative texts (in our experiments we find $1000$ to be large enough; while a larger number can marginally improve the final performance but it incurs a larger computation burden), then we compute the dot products of the current image feature and all these sampled text features, and finally, we rank the dot products and select the top $N - 1$ negatives to be included in computing the gradients of the loss (we find $N=8$ to be effective).
We use the asymmetric version of the cross-entropy loss (i.e., the normal one) compared to the asymmetric version in CLIP pre-training since the number of negatives per image does not equal the batch size when HNM is adopted.
We reduce the batch size to 4 whenever with online HNM so that directly fine-tuning the largest CLIP model is viable with a single V100 Nvidia GPU.
We still apply the learnable ``logit scale'' parameter in CLIP pre-training which effectively makes contrastive learning more stable.

For full HNM, if we directly fine-tune the CLIP model, we need to compute text features of all texts in the training set in every gradient step.
While this is computationally prohibitive, we adopt the feature adapter strategy and thus cache all the text features once and do not update the text encoder and the text features during fine-tuning.

\subsection{More Ablation Studies}
In our experiments presented in the main paper (specifically in Tab. 1), we have justified the use of online HNM, the additional inputs (scene-text and brand information) to the feature adaptation, and the advantages of the attention-based adapter over the baseline adapter.
We also perform experiments on several variants of the attention-based feature adapter and find that either using more than one attention layer or adding layer norm \& additional linear projection as in the encoder-decoder Transformer \cite{vaswani2017attention} make the model more vulnerable to overfitting.

\subsection{Additional Details of Feature Adapters}
For feature adapters (CLIP-Adapter and KAFA), we use the full HNM for fine-tuning as discussed in the previous section.
We use the same training setup as that of ``Direct ft + HMN'' except for the additional input branches.
For CLIP-Adapter, we tailor it to our setup by training three 2-layer residual MLPs.
Specifically, let as denote them as $g^{mlp}_I$, $g^{mlp}_T$ and $h^{mlp}$, built on top of the image and text features extracted by VLMs, and a mixture of these features, respectively.
The adapted feature for $\textbf{x}$ becomes
\begin{align*}
    f^{mlp}_{I}(x) &= \texttt{n}(f_{I}(x) + g^{mlp}_{I}(f_{I}(x))) \\
    f^{mlp}_{T}(x_{st}) &= \texttt{n}(f_{T}(x_{st}) + g^{mlp}_{T}(f_{T}(x_{st}))) \\
    f^{mlp}(x) &= \texttt{n}(h^{mlp}(\textrm{cat}[f^{mlp}_{I}(x), f^{mlp}_{T}(x_{st}), ...])) 
\end{align*}
where $\texttt{cat}$ is concatenation. Here we omit the adapted feature for text label $y$.
And the adapted feature for the text label $y$ becomes
\begin{equation*}
    f^{mlp}_{T}(y) = \texttt{n}(f_{T}(y) + g^{mlp}_{T}(f_{T}(y))) \\
\end{equation*}
which is used during full HNM for fine-tuning.

For fine-tuning of both CLIP-Adapter and KAFA, we find a much larger learning rate (i.e., $1e-4$) to be effective and train the model similarly with early stopping and a maximum of 10 epochs.
We find it helpful to stabilize training by adding an additional regularization loss to keep the feature adapter’s output close to the VLM image features.
Specifically, we add the negative of dot products between the two (averaged over all data points in the mini-batch) to the overall training objective.
For this regularization term, we use a coefficient of $5$ in all our experiments in the Pitt Dataset.

\section{Additional Examples} \label{sec:more_examples}
We present 2 additional examples in Fig. \ref{fig:more_examples} to illustrate the improvement of our method over the baseline.
Again, we only show 2 negative text descriptions for better display, and we anonymize all brand info.

\end{document}